%% file: Topo-LSTM.tex
\documentclass[conference]{IEEEtran}
\pdfoutput=1

\usepackage{graphicx}
\usepackage{balance}  
\usepackage{url}
\usepackage{amssymb}
\usepackage{amsmath}
\usepackage{algorithm}
\usepackage{epstopdf}
\usepackage{algorithmic}

\newtheorem{definition}{Definition}

\newcommand{\stitle}[1]{\vspace{2mm} \noindent {\bf #1}}
\newcommand{\eat}[1]{}
\newcommand{\eg}{{\it e.g.}}
\newcommand{\vs}{{\it v.s.}}

\newcommand{\ie}{{\it i.e.}}

\newcommand{\wrt}{w.r.t. }

\newcommand{\boldx}{\mathbf{x}}
\newcommand{\boldh}{\mathbf{h}}
\newcommand{\boldg}{\mathbf{g}}
\newcommand{\boldi}{\mathbf{i}}
\newcommand{\boldf}{\mathbf{f}}
\newcommand{\boldo}{\mathbf{o}}
\newcommand{\boldc}{\mathbf{c}}

\newcommand{\inW}{W_i}
\newcommand{\inU}{U_i}
\newcommand{\inb}{\mathbf{b}_i}
\newcommand{\outW}{W_o}
\newcommand{\outU}{U_o}
\newcommand{\outb}{\mathbf{b}_o}
\newcommand{\fgetW}{W_f}

\newcommand{\fgetUp}{U_{fp}}
\newcommand{\fgetUq}{U_{fq}}
\newcommand{\fgetb}{\mathbf{b}_f}
\newcommand{\cellW}{W_c}
\newcommand{\cellU}{U_c}
\newcommand{\cellb}{\mathbf{b}_c}

\begin{document}
%
\title{Topological Recurrent Neural Network for Diffusion Prediction}

\author{\makebox[0.5\linewidth]{Jia Wang}\\
University of Illinois at Urbana-Champaign\\
cnwangjia@gmail.com\\
\and
\makebox[0.5\linewidth]{Vincent W. Zheng}\\
Advanced Digital Sciences Center, Singapore\\
vincent.zheng@adsc.com.sg\\
\and
\makebox[0.5\linewidth]{Zemin Liu}\\
Zhejiang University\\
liuzemin@zju.edu.cn
\and
\makebox[0.5\linewidth]{Kevin Chen-Chuan Chang}\\
University of Illinois at Urbana-Champaign\\
kcchang@illinois.edu
}


%


\maketitle

\begin{abstract}
In this paper, we study the problem of using representation learning to assist information diffusion prediction on graphs. In particular, we aim at estimating the probability of an inactive node to be activated next in a cascade. Despite the success of recent deep learning methods for diffusion, we find that they often underexplore the cascade structure. We consider a cascade as not merely a sequence of nodes ordered by their activation time stamps; instead, it has a richer structure indicating the diffusion process over the data graph. As a result, we introduce a new data model, namely diffusion topologies, to fully describe the cascade structure. We find it challenging to model diffusion topologies, which are dynamic directed acyclic graphs (DAGs), with the existing neural networks. Therefore, we propose a novel topological recurrent neural network, namely Topo-LSTM, for modeling dynamic DAGs. We customize Topo-LSTM for the diffusion prediction task, and show it improves the state-of-the-art baselines, by 20.1\%--56.6\% (MAP) relatively, across multiple real-world data sets. Our code and data sets are available online\footnote{\url{https://github.com/vwz/topolstm}}.
 
\end{abstract}


%
\IEEEpeerreviewmaketitle

\input{intro}

\input{related_work}

\input{technical_section}

\input{experiment_section}

\input{conclusion}


\section*{Acknowledgment}
We thank the support of: National Natural Science Foundation of China (No. 61502418), Research Grant for Human-centered Cyber-physical Systems Programme at Advanced Digital Sciences Center from Singapore A*STAR, and the National Science Foundation Grant No. IIS 16-19302. Any opinions, findings, and conclusions or recommendations expressed in this publication are those of the author(s) and do not necessarily reflect the views of the funding agencies.

\bibliographystyle{IEEEtran}
\bibliography{ref}

\end{document}

%% file: intro.tex
\section{Introduction} \label{sect.intro}

\begin{figure}[t]
\includegraphics[width=1\linewidth]{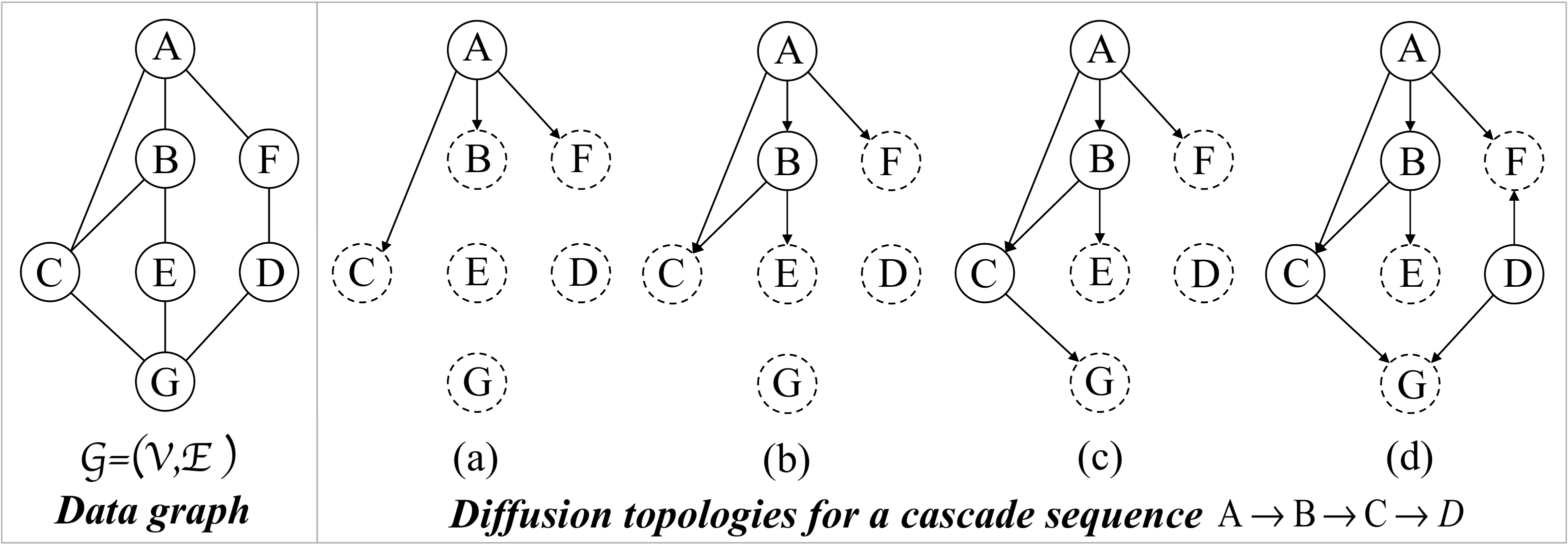}
\caption{Illustration of diffusion cascade modeling. At each time stamp $t$, we construct a diffusion topology, which characterizes the cascade status until $t$.  In each diffusion topology, a solid circle is an \emph{active} node, a dotted circle is an \emph{inactive} node, and an arrow is a possible activation attempt.}
\label{fig:diffusion_topology}
\end{figure}

Information diffusion is a common phenomenon on social networks \cite{DBLP:conf/icwsm/LermanG10,DBLP:conf/aaai/ZhangTZMLSHS17}. 
Its modeling has many applications, such as helping to predict
which user is an opinion leader \cite{DBLP:conf/kdd/KempeKT03}, 
how much a cascade will grow \cite{DBLP:conf/www/ChengADKL14}, 
who are the diffusion sources \cite{DBLP:conf/aaai/ZhuCY17}, 
which user will digg a particular story \cite{DBLP:conf/ijcai/BaoCL16}, and so on.
In this paper, we study the task of information diffusion prediction. The goal is to design an effective diffusion model, which can estimate the activation probability for an inactive node in a cascade. 
We consider the most standard setting of information diffusion, where we have inputs of: 1) a \emph{data graph} $\mathcal{G} = (\mathcal{V}, \mathcal{E})$, where $\mathcal{V}$ is the set of nodes and $\mathcal{E}$ is the set of edges; 2) a set of \emph{cascade sequences}, each of which is an ordered sequence of node activation over $\mathcal{V}$. For example, in Fig.~\ref{fig:diffusion_topology}, the data graph $\mathcal{G}$ is a network of seven nodes; a cascade sequence $A \!\rightarrow\! B  \!\rightarrow\! C \!\rightarrow\! D$ is a sequence of nodes ordered by their activation time stamps. 

Early work assumes diffusion model as given, such as independent cascade (IC) and linear threshold (LT) \cite{DBLP:conf/kdd/KempeKT03}. There are many extensions of the IC and LT models, such as continuous-time IC \cite{DBLP:conf/icml/Gomez-RodriguezBS11}. Besides, the IC and LT models also enable an important research direction of influence maximization~\cite{DBLP:conf/kdd/Gomez-RodriguezLK10,DBLP:conf/soda/BorgsBCL14,DBLP:conf/sigmod/YangMPH16}.
Recent work tries to learn a diffusion model from the available cascade data. They often rely on explicitly engineering useful features to predict the activation probability of a node, such as network structure and temporal information~\cite{DBLP:conf/www/ChengADKL14}, user nodes' social roles~\cite{DBLP:conf/aaai/YangTLSCLY15}, diffusion content~\cite{DBLP:conf/wsdm/TsurR12} and user nodes' interactions~\cite{DBLP:conf/wsdm/GoyalBL10}. Although these methods have shown significant improvements in diffusion prediction performance, the feature engineering process requires much manual effort and extensive domain knowledge. 
With the recent development of neural networks, recent work starts to exploit deep learning, so as to avoid explicit feature engineering for diffusion modeling. A small number of pioneer work uses graph embedding to model diffusion. For example, Embedded-IC~\cite{DBLP:conf/wsdm/BourigaultLG16} takes a cascade-based modeling approach, which considers each inactive node to be activated by the active nodes. It differentiates two kinds of roles for the nodes; \ie, an active node serves as a ``sender'', and an inactive node serves as a ``receiver'', so that the inactive node receives information from the active nodes in a diffusion cascade. For each role, it learns a vector as a node's embedding. Then, it models an activation based on the closeness between an inactive node's receiver embedding vector and the active nodes' sender embedding vectors. DeepCas~\cite{DBLP:conf/www/LiMGM17} is designed to predict the future cascade size. It models the cascade at each time step with an induced subgraph over the active nodes. Then, it decomposes the subgraph into some random walk paths, and uses Gated Recurrent Unit (GRU)~\cite{DBLP:journals/corr/ChungGCB14} to learn an embedding vector of the subgraph. Based on this subgraph embedding vector, it predicts the cascade size in the future. 

Despite the success of these deep learning methods for diffusion modeling, we find that they often underexplore the cascade structure. A cascade is not merely a sequence of nodes ordered by the activation time stamps; instead, it has a richer structure indicating the diffusion dynamics over the data graph. For example, in Fig.~\ref{fig:diffusion_topology}, to represent the cascade from $A$ to $D$, we shall describe how it spreads over the data graph $\mathcal{G}$. Particularly, when $A$ is activated at the first place, it has a chance to activate its neighbors in $\mathcal{G}$, which are $\{B, C, F\}$. By drawing an arrow from $A$ to each of its neighbors to show the possible activation attempts, we have a \emph{diffusion topology} to characterize the cascade until the current time stamp in Fig.~\ref{fig:diffusion_topology}(a). Similarly, when $B$ is activated next, it also has a chance to activate its (inactive) neighbors $\{C, E\}$. As a result, we have another diffusion topology in Fig.~\ref{fig:diffusion_topology}(b) for the new time stamp. 
Such diffusion topologies are useful for diffusion prediction; \eg, a Twitter user is more likely to propagate a piece of news to her friends, if that news has been retweeted by many celebrities. However, the existing deep learning methods for diffusion modeling do not take  diffusion topologies into consideration. For example, in Embedded-IC \cite{DBLP:conf/wsdm/BourigaultLG16}, an activation at time stamp $t$ is enabled between an inactive node and all the existing active nodes by $t$, regardless of the network structure. Therefore, the embedding of each sender (\ie, active node) is learned without the diffusion topologies. In DeepCas \cite{DBLP:conf/www/LiMGM17}, the induced subgraph at each time step does not capture how the diffusion spreads; besides, each subgraph is further decomposed into paths and embedded independently, thus the resulting node embedding is only partially aware of the data graph structure. 

In this paper, we study how to fully explore the cascade structure by deep learning for diffusion prediction. Generally, in cascade dynamics, the active nodes try to ``send'' information to the inactive nodes; upon successfully ``receiving'' the information, an inactive node becomes activated. Motivated by such cascade dynamics, we choose to differentiate two roles for each node; \ie, each active node acts as a ``sender'' and each inactive node acts as a ``receiver''. Therefore, to enable deep learning for diffusion modeling, we try to embed each active node with a sender embedding vector, and each inactive node with a receiver embedding vector, such that we can simply predict a node activation based on these embedding vectors. Although such a ``sender''-vs-``receiver'' role differentiation is also adopted by Embedded-IC, we approach the embedding problem differently. In Embedded-IC, each sender (or receiver) embedding is considered as encoding the \emph{static preferences} of a sender (or receiver). Such a formulation overlooks the \emph{dynamic context} of diffusion topologies; \ie, once the sender embedding of an active node and the receiver embedding of an inactive node are fixed, the resulting activation probability based on these two embedding vectors is also fixed, regardless of how the cascade grows over time. As discussed earlier, such dynamic context of diffusion topologies are useful for diffusion prediction. Therefore, we should consider the sender embedding as encoding not only the active node's static tendency, but also the dynamic context of the diffusion topology. As the inactive nodes have not participated in the cascade so far, it is reasonable to consider each receiver embedding as only encoding its inactive node's static preferences. 

Technically, it is not trivial to learn sender embedding with diffusion topologies. This is because each diffusion topology is a directed acyclic graph (DAG) and it evolves over time. On the one hand, we cannot over-simplify the dynamic DAGs into a set of independent nodes or a set of random walk paths for learning the sender/receiver embedding, since these over-simplified formulations are unable to fully exploit the topologies of cascades on the network. On the other hand, due to the dynamic nature of diffusion topologies, we are looking for a recurrent neural network (RNN) formulation. To the best of our knowledge, however, there is no RNN that is able to handle such dynamic DAG structure of a diffusion. For example, the existing RNN models mainly focus on either sequence-structured inputs, such as Long Short-Term Memory (LSTM)~\cite{HochreiterS97} and GRU, or tree-structured inputs, such as Tree-LSTM~\cite{DBLP:conf/acl/TaiSM15}. There are a handful of RNN models that try to model static DAGs designed for different application domains; \eg, DAG-RNN~\cite{DBLP:conf/cvpr/ShuaiZWW16,DBLP:journals/jmlr/BaldiP03} models each 2D image as a DAG for scene labeling, while RNN-LE~\cite{DBLP:journals/nn/BianchiniMSS05} models each contact map over a protein's amino acids as a DAG for protein structure prediction. However, both DAG-RNN and RNN-LE are based on the plain RNN architecture, and are unable to capture the peculiarities of a diffusion process. Thus, an RNN architecture tailed for diffusion is required.

To model the diffusion topologies, we propose a novel Topological LSTM (Topo-LSTM) model. 
Topo-LSTM is a DAG-structured RNN, which takes dynamic DAGs as inputs and generates a topology-aware embedding for each node in the DAGs as outputs. In the application of diffusion prediction, we use Topo-LSTM to learn the sender embedding for each node $v_t$ activated at time $t$ in a cascade. We ensure the learned sender embedding of $v_t$ as fully aware of which other nodes have been activated so far and how the diffusion spreads to reach $v_t$. For example, in Fig.~\ref{fig:diffusion_topology}, the sender embedding of an active node $C$ knows that $A$ and $B$ have been activated, and the diffusion spreads like Fig.~\ref{fig:diffusion_topology}(b) before activating $C$. We consider each inactive node as having a receiver embedding, which is independent of the cascade to indicate the node's intrinsic preference. We also learn the receiver embedding for each inactive node, and use it to predict an activation based on its closeness to the active nodes' sender embeddings. It is worth noting that, our model has few hyperparameters, including an embedding dimension and a trade-off parameter for model regularization. This makes our model easy to tune in practice, compared with other graph-based deep learning methods, which require additional hyperparameters for either graph sampling or objective functions~\cite{perozzi2014deepwalk,GroverL16,DBLP:conf/www/LiMGM17}. 

We summarize our contributions as follows. 

\begin{itemize}
\item We propose a new data model, namely diffusion topology, to fully explore the diffusion structure.
\item We propose a novel Topo-LSTM model, which is able to handle the dynamic DAG structure of diffusion topologies and tailored for the task of node activation prediction. 
\item We evaluate Topo-LSTM on several public real-world data sets, and show that Topo-LSTM significantly outperforms the state-of-the-art baselines.
\end{itemize}

%% file: related_work.tex
\section{Related Work} \label{sect.related_work}


In diffusion prediction, early work assumes the diffusion model as given. 
For example, in \cite{DBLP:conf/kdd/KempeKT03}, IC and LT models are used for influence maximization.  
Some other work tries to learn the diffusion model from the cascade data. 
For example, in \cite{DBLP:conf/icml/DuLBS14}, the diffusion model is formulated as a learnable coverage function. In \cite{DBLP:conf/kdd/MyersZL12}, both the internal and external influences are modeled, with some parameters to be learned. 
In this paper, we focus on the task of diffusion prediction, and aim to learn an activation function from the data. 
In diffusion prediction, most existing work relies on extracting useful features for activation prediction. 
For example, in \cite{DBLP:conf/www/ChengADKL14}, various features are exploited, including content, user, time and network structure. Some recent studies use deep learning to avoid feature engineering for diffusion prediction. 
As introduced in Sect.~\ref{sect.intro}, both Embedded-IC \cite{DBLP:conf/wsdm/BourigaultLG16} and DeepCas \cite{DBLP:conf/www/LiMGM17} are shown to significantly improve the diffusion prediction. Wang et al.~\cite{wang2015learning} proposes a model similar to Embedded-IC which computes two low-dimensional vectors for each node to capture its influence and susceptibility. However, they tend to underexplore the cascade structure. In comparison, we try to fully explore the diffusion dynamics with diffusion topologies. Recently, Du et al.~\cite{du2016recurrent} uses RNN to model linear sequences of events with timestamps. However, their model cannot be applied to information diffusion in networks.

In the recent development of recurrent and recursive neural networks, multiple types of data structures are considered. 
Standard RNNs are designed for sequences; \eg, LSTM and GRU are used to model music and speech \cite{DBLP:journals/corr/ChungGCB14}. 
Tree-structured RNN are exploited especially in natural language processing. For example, to model the dependency tree in word embedding, a DT-RNN is proposed \cite{DBLP:journals/tacl/SocherKLMN14}. In \cite{DBLP:conf/acl/TaiSM15}, a tree-LSTM is introduced to model the syntactic properties of combining words to phrases. In \cite{DBLP:conf/naacl/DyerKBS16}, a RNNG is developed to encode a parse tree. 
There are a handful of RNNs designed for DAGs in various applications. 
For example, in scene labeling \cite{DBLP:conf/cvpr/ShuaiZWW16}, an image is segmented into a 2D lattice, from which several DAGs are extracted by a tree-reweighted max-product algorithm; then, a DAG-RNN is introduced for modeling. 
In protein structure prediction \cite{DBLP:journals/jmlr/BaldiP03}, a \emph{contact map} over the amino acids of a protein is decomposed into a DAG by traversing the map in certain directions; then another DAG-RNN is developed. 
In face detection from images \cite{DBLP:journals/nn/BianchiniMSS05}, a \emph{region adjacency graph with labeled edges} is first extracted from each image; then, the graph is decomposed into an \emph{edge-labeled DAG} by breadth-first search, and modeled by a RNN-LE model.

Compared with the above DAG-structured RNNs, we have two major differences. 
First, our Topo-LSTM is designed for dynamic DAGs, where the DAGs (\ie, diffusion topologies) evolve over time. 
In contrast, the input DAGs of both DAG-RNNs and RNN-LE are static (\eg, images and protein structures). 
Second, our Topo-LSTM is designed for a different application. 
Due to domain differences, we cannot directly apply DAG-RNNs and RNN-LE to our diffusion prediction task. 
For example, DAG-RNNs are customized for images, which are 2D lattices with fixed orders, instead of a real graph as used in information diffusion. RNN-LE requires the DAGs to have edge labels, which are not available in our problem. Also, at each time step, in their models a recurrent unit can take only take a single type of inputs from their precedents, whereas the diffusion problem requires our model to take inputs from different types of precedents (\ie, predecessors of the current node \vs others in the diffusion topology), in order to account for the different types of influences from previously activated nodes on the current node. Third, they are based on the plain RNN architecture, which are insufficient to model the complexity of a diffusion process, and suffer from the vanishing/exploding gradient issues.

Finally, there is a relevant line of research on \emph{graph embedding}. Most graph embedding aims to output a vector representation for each node in the graph, such that two nodes ``close'' on the graph have similar vector representations in a low-dimensional space. Specifically, earlier node embedding methods, such as LLE \cite{RoweisS00}, often focus on preserving \emph{first-order} proximity, where two nodes directly linked in the graphs have similar embedding vectors. More recent methods, such as DeepWalk \cite{perozzi2014deepwalk} and Node2Vec \cite{GroverL16}, start to consider preserving \emph{second-order} proximity, where two nodes sharing similar ``neighbors'' have similar embedding vectors. LINE \cite{TangQWZYM15} tries to preserve both first- and second-order proximity. GraRep \cite{CaoLX15} and HOPE \cite{OuCPZZ16} preserve high-order proximity by learning node embedding from a high-order adjacency matrix. ComE \cite{DBLP:journals/corr/ZhengCCCC16,DBLP:conf/cikm/CavallariZCCC17} further considers community-aware high-order proximity for node embedding and community embedding. There is also some graph embedding work that focuses on either edge embedding (\eg, TransE \cite{DBLP:conf/nips/BordesUGWY13}), path embedding (\eg, ProxEmbed \cite{DBLP:conf/aaai/LiuZZZCWY17}), structure embedding (\eg, topoLSTM \cite{WangZLC17}) or whole graph embedding (\eg, Graph Convolutional Network \cite{DBLP:journals/cilr/KipfW17}). A comprehensive survey of graph embedding with different types of outputs is in \cite{DBLP:journals/corr/abs-1709-07604}. 
Although the above methods are successful for many applications, they are designed for general purpose graph embedding to preserve the data graph structure, thus it is unable to incorporate the cascade information for diffusion modeling.

%% file: technical_section.tex
\section{Diffusion Modeling}

We study the task of diffusion prediction that aims to estimate which node to activate next based on a cascade of node activations on a data graph. To formalize our problem, we first introduce some terminologies. We also summarize our notations in Table~\ref{tab:notation} for reader's reference.

\begin{table}[t]
	\centering
	\caption{List of notations.}
	\label{tab:notation}
	\begin{tabular}{c|c}
		\hline
		\textbf{Symbol}      & \textbf{Description}                         \\ \hline
		$v_t$                & The node activated at time $t$ in a cascade. \\
		$\boldh_t$           & Sender embedding of active node $v_t$.       \\
		$\mathbf{g}_j$       & Receiver embedding of inactive node $v_j$.   \\
		$Q_{1:t-1}$          & Set of active nodes before time $t$          \\
		$\mathcal{G}^*_t$    & Diffusion topology at time $t$.              \\
		$\mathcal{P}_{v, t}$ & Precedent set of node $v$ at time $t$.       \\
		$\boldx_v$           & Feature vector of node $v$.                  \\
		$\phi(\cdot)$        & An aggregation operator.                     \\
		$\rho_{v,t}$         & Activation score of node $v$ at time $t$.    \\
		\hline
	\end{tabular}
\end{table}

\begin{definition}
A \textbf{data graph} is $\mathcal{G} = (\mathcal{V}, \mathcal{E})$, where $\mathcal{V}$ is the node set, and $\mathcal{E}$ is the edge set.  
\end{definition}

\begin{definition}
A \textbf{cascade sequence} is an ordered sequence of tuples $s = \{(v_1, t_1), ..., (v_T, t_T)\}$, where each $v_j$ is a distinct node in $\mathcal{V}$, and each $t_j$ is a time stamp such that $t_j < t_{j+1}$. 
\end{definition}

As \emph{problem input}, we have a data graph $\mathcal{G}$ and a set of training cascade sequences $\mathcal{S} = \{s_1, ..., s_n\}$. It is worth noting that, in this paper we consider the most basic setting of information diffusion \cite{DBLP:conf/wsdm/BourigaultLG16,DBLP:conf/www/LiMGM17}. Specifically, we do not assume the diffusion content information as available in the data graph and the cascade sequences. Besides, we do not assume the exact time information (\ie, what time each $t_j$ refers to) in each cascade sequence as known either; instead, we only rely on the order of the nodes in each cascade sequence, and thus we could equivalently rewrite the cascade as $s=\{(v_1, 1), ..., (v_T, T)\}$. We leave incorporating content and exact time information as future work. As \emph{problem output}, we have a diffusion model $\mathcal{M}$, which is able to predict a node to activate at time $t$, given a test cascade sequence $s' = \{(v_1', 1), ..., (v_{t-1}', t-1) \}$. 

We use a deep learning approach to learn $\mathcal{M}$. 
As motivated in Sect.~\ref{sect.intro}, we consider the ``sender''-vs-``receiver'' role differentiation in node embedding. Particularly, we try to learn a \emph{sender embedding} $\boldh_t \in \mathbb{R}^d$ for each active node $v_t$, and a \emph{receiver embedding} $\mathbf{g}_j \in \mathbb{R}^d$ for each inactive node $v_j$. As a result, we define the activation probability of an inactive node $v_j$ at time $t$ by a function over $\boldg_j$ and all the active nodes $\{ \boldh_1, ..., \boldh_t \}$. Ideally, we expect the sender embedding $\boldh_t$ to encode not only the static preferences of $v_t$, but also the dynamic contex of how a cascade propagates to $v_t$ until time $t$. That is, we need to learn $\boldh_t$ progressively over time for each cascade sequence, based on who $v_t$ is and how the cascade dynamics looks like.  As the inactive node $v_j$ has not participate in the cascade yet, we let its receiver embedding $\mathbf{g}_j$ only encode the static preferences of $v_j$. That is, $\mathbf{g}_j$ is only dependent on who $v_j$ is, regardless of the cascade sequences. 

The challenge of learning $\mathcal{M}$ is that, we need to make the sender embedding be fully aware of the cascade dynamics, which describes how a cascade sequence spreads over the data graph. To address this challenge, we introduce a new data model, namely diffusion topology, to model the cascades (Sect.~\ref{sect.data_model}). Then we develop a novel Topo-LSTM model to learn the sender embedding for active nodes and the receiver embedding for inactive node with the diffusion topologies (Sect.~\ref{sect.dt_embedding}). Finally, we use both the sender embeddings and the receiver embeddings to develop an activation function, and use the ground truth node activation as supervision to train the Topo-LSTM (Sect.~\ref{sect.activation_pred}). 

\subsection{Diffusion Topology as Data Model} \label{sect.data_model} 

We discuss how to prepare the cascade data for learning the sender embedding for each active node. 
For a cascade sequence $s = \{(v_1, 1)$, ..., $(v_T, T)\}$, we denote $Q_{1:t-1}$ as the set of active nodes in $s$ before time $t$; \ie, $Q_{1:t-1} = \{v_1, ..., v_{t-1}\}$. Ideally, as the sender embedding of $v_t$, $\boldh_t$ needs to be fully aware of the cascade dynamics; \ie, it knows not only which nodes are in $Q_{1:t-1}$, but also how the diffusion spreads to reach $v_t$. Let us consider the cascade sequence in Fig.~\ref{fig:diffusion_topology}. $C$'s sender embedding $\boldh_C$ is supposed to encode the cascade dynamics that, $Q_{1:2} = \{A, B\}$ have been activated and the diffusion, before activating $C$, has spreaded like Fig.~\ref{fig:diffusion_topology}(b). In general, a diffusion topology such as Fig.~\ref{fig:diffusion_topology}(d) is not explicitly available in $Q_{1:t-1}$; instead, it needs to be constructed from $Q_{1:t-1}$ and the data graph $\mathcal{G}$. We remark that, for each $v_t$, there is only one unique diffusion topology; this is because at different time stamps, the set of active nodes and their cascade structures are different. 

\begin{definition} \label{def.dt}
For a cascade $s = \{(v_1, 1), ..., (v_T, T)\}$ and a data graph $\mathcal{G} = (\mathcal{V}, \mathcal{E})$, the \textbf{diffusion topology} of $s$ at time $t$ is a directed graph $\mathcal{G}^*_t = (\mathcal{V}, \mathcal{E}^*_t)$, where $\mathcal{E}^*_t = \{ (v_i, u) | (v_i, u) \in \mathcal{E}, v_i \in Q_{1:t-1}, u \in (\mathcal{V} \backslash Q_{1:t-1}) \cup Q_{i+1:t-1} \}$ is a set of directed edges, indicating all the possible activation attempts until $t$.
\end{definition}

For an edge $(v_i, u) \in \mathcal{E}^*_t$, $v_i$ already became active at time $i \leq t-1$. Depending on whether $u$ is active or not by the time $t-1$, this edge $(v_i, u)$ has different kinds of semantics. Specifically, if $u \in (\mathcal{V} \backslash Q_{1:t-1})$, \ie, $u$ is inactive by time $t-1$, then $(v_i, u)$ indicates a possible ``future activation'' attempt from $v_i$ to $u$. If $u \in Q_{i+1:t-1}$, \ie, $u$ became active after time $i$ and before time $t-1$, then it indicates a possible ``past activation'' from $v_i$ to $u$. Because in both of the above cases, $v_i$ always tries to activate $u$, we call $v_i$ as a \emph{precedent} of $u$, for each $(v_i, u) \in \mathcal{E}^*_t$. Given the diffusion topology $\mathcal{G}^*_t = (\mathcal{V}, \mathcal{E}^*_t)$ at time $t$, we denote the precedent set of each $v\in \mathcal{V}$ at time $t$ as $\mathcal{P}_{v, t} = \{ v_i | (v_i, v) \in \mathcal{E}^*_t \}$. 
As we can see, a diffusion topology fully characterizes the cascade structure on the data graph, thus it is suitable for us to use as a data model for diffusion representation learning. 

\stitle{Properties and implications.} 
We conclude two important properties of diffusion topology from Def.~\ref{def.dt}:

\begin{itemize}
\item Each diffusion topology is a \emph{directed acyclic graph} (DAG). This is because the directed edges in a diffusion topology are always from a node activated earlier, to another node which is to be activated later; there is strictly no cycle. 
\item The diffusion topologies for a cascade are \emph{monotonically growing} over time. This is because the diffusion topology at time $t$ is always a supergraph of that at time $t-1$, due to its introducing new edges. 
\end{itemize}

The two properties of diffusion topology has important implications.  
Firstly, learning $\boldh_t$ with dynamic diffusion topologies is not trivial, because no prior RNNs are designed for dynamic DAGs. This motivates us to design a novel neural network model. 
Secondly, $\boldh_t$ can be learned \emph{recurrently} from the earlier $\boldh_i$'s $(i=1, ..., t-1)$, since these $\boldh_i$'s have encoded the diffusion topologies before time $t$, which are essentially subgraphs of the diffusion topology at time $t$. 

In all, we propose to use diffusion topologies as our data model to learn the sender embedding. We emphasize that our diffusion topology data model is new. In the existing literature, Embedded-IC's data model is a set of independent nodes \cite{DBLP:conf/wsdm/BourigaultLG16}, which are not aware of the data graph structure; DeepCas's data model is a set of independent paths sampled from the induced subgraph over the active nodes at each time $t$ \cite{DBLP:conf/www/LiMGM17}, and each path alone only partially captures the cascade structure. 

\subsection{Diffusion Topology Embedding} \label{sect.dt_embedding} 

Next we discuss how to learn $v_t$'s sender embedding $\boldh_t$ from the diffusion topology $\mathcal{G}^*_t$ at time $t$ and the earlier activated nodes $Q_{1:t-1}$'s sender embeddings $\{ \boldh_1, ... , \boldh_{t-1} \}$. As discussed in Sect.~\ref{sect.data_model}, $\boldh_t$ can be learned recurrently from $\{ \boldh_1, ... , \boldh_{t-1} \}$. This motivates us to extend a Recurrent Neural Network (RNN) framework to develop our embedding model. 

LSTM is a popular neural network architecture designed for RNN to address the vanishing/exploding gradient issues \cite{HochreiterS97}. The unit of an LSTM network is the \emph{memory cell}, which has an input gate, a neuron with a self-recurrent connection, a forget gate, and an output gate. The input gate allows incoming signal to alter the memory cell's state or block it. The self-recurrent connection balances signals from the previous time step and the current time step. The forget gate modulates the memory cell's self-recurrent connection, allowing the cell to remember or forget its previous state, as needed. The output gate allows the state of the memory cell to affect other neurons or prevent it. The standard LSTM is designed for sequences, but not DAGs. The recent Tree-LSTM \cite{DBLP:conf/acl/TaiSM15} cannot handle DAGs either. The existing RNN models that take DAGs as inputs, such as DAG-RNN  \cite{DBLP:conf/cvpr/ShuaiZWW16} and RNN-LE \cite{DBLP:journals/nn/BianchiniMSS05}, do not exploit the LSTM architecture. To the best of our knowledge, LSTM has not been used to model DAGs before. Besides, these existing RNN architectures are designed for different application domains and are not applicable to our problem.

\begin{figure*}[t]
	\centering
	\includegraphics[width=0.6\linewidth]{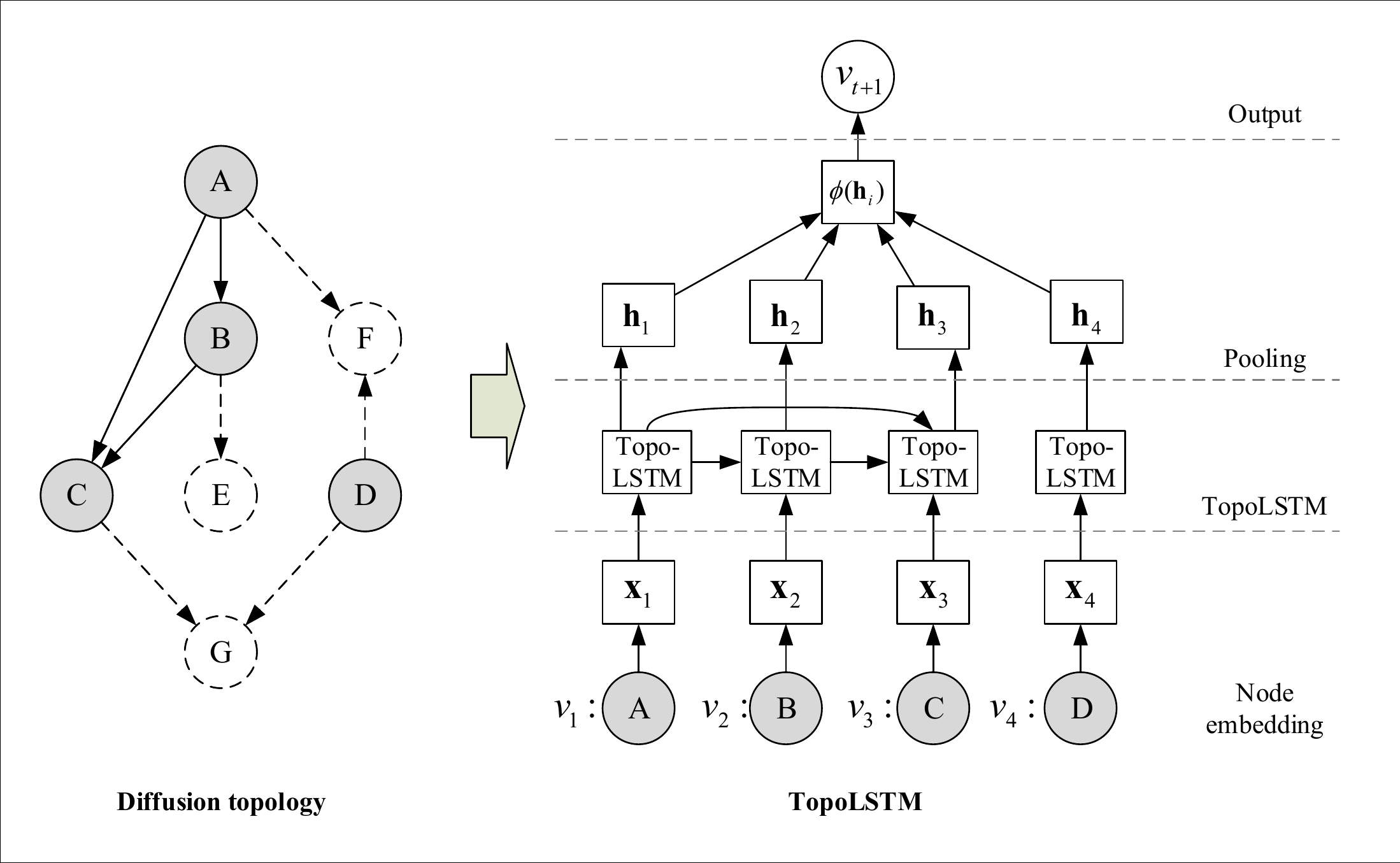}
	\caption{The Topo-LSTM framework (using the diffusion topology in Fig~\ref{fig:diffusion_topology}(d) as an example).}
	\label{fig:frmk}
\end{figure*}

We extend standard LSTM to Topo-LSTM for modeling the diffusion topologies, which are DAGs. 
To assist model development, we use Fig.~\ref{fig:diffusion_topology} as a running example. The overall architecture of the Topo-LSTM model is also illustrated in Fig.~\ref{fig:frmk} using the diffusion topology in Fig~\ref{fig:diffusion_topology}(d) as an example.

\stitle{Running Example.} 
As in Fig.~\ref{fig:diffusion_topology}, we are given a cascade sequence $\{(A, 1), (B, 2), (C, 3), (D, 4)\}$. At time $t=1$, the diffusion topology is $\mathcal{G}^*_1 = (\mathcal{V}, \mathcal{E}^*_1)$ with $\mathcal{E}^*_1 = \emptyset$. Denote $\boldx_i \in \{0, 1\}^{|\mathcal{V}|}$ as the feature vector for node $v_i$. For example, $\boldx_i$ could be the \emph{one-hot} ID vector where $\boldx_i$ has 1 on its $i$th entry and all zeros elsewhere. As illustrated in the network slice at $t=1$ in Fig.~\ref{fig:frmk}, similar to standard LSTM, we take $A$'s feature vector $\boldx_A$ as input and transform it to a dense vector $\boldh_A$ as output. At $t=2$, as $A$ has been activated, we have $Q_{1:1} = \{ A \}$ and $\mathcal{G}^*_2 = (\mathcal{V}, \mathcal{E}^*_2)$ with $\mathcal{E}^*_2 = \{ (A,B), (A,C), (A,F) \}$, as shown in Fig.~\ref{fig:diffusion_topology}(a). The precedent set for $B$ is $\mathcal{P}_{B,2} = \{A\}$. To make $\boldh_B$ aware of the cascade structure so far, we infer $\boldh_B$ from both $B$'s feature vector $\boldx_B$ and the possible activation attempt from $\mathcal{P}_{B,2}$. These operations at $t=2$ are illustrated by the network slice at $t=2$ in Fig.~\ref{fig:frmk}. At $t=3$, we have $Q_{1:2} = \{A, B\}$ and $\mathcal{G}^*_3 = (\mathcal{V}, \mathcal{E}^*_3)$ with $\mathcal{E}^*_3 = \{ (A,B), (A,C), (A,F), (B,C), (B,E) \}$, as shown in Fig.~\ref{fig:diffusion_topology}(b). Given $\mathcal{P}_{C,3} = \{A, B \}$, we infer $\boldh_C$ from $C$'s feature vector $\boldx_C$ and the possible activation attempts from $\mathcal{P}_{C,3}$. The network slice at $t=3$ in Fig.~\ref{fig:frmk} depicts the above operations for $t=3$.  At $t=4$, $Q_{1:3} = \{A, B, C\}$ but this time there is no link from $Q_{1:3}$ to $D$ on $\mathcal{G}$. As also shown in the slice at $t=4$ in Fig.~\ref{fig:frmk}, there is no incoming edge into Topo-LSTM unit corresponding to $t=4$. Thus $\mathcal{P}_{D,4} = \emptyset$ and $\mathcal{G}^*_4 = (\mathcal{V}, \mathcal{E}^*_4)$ with $\mathcal{E}^*_4 = \{ (A,B), (A,C), (A,F), (B,C), (B,E), (C,G) \}$, as shown in Fig.~\ref{fig:diffusion_topology}(c). Then, we infer $\boldh_D$ from $D$'s feature vector $\boldx_D$ and the other already activated nodes $Q_{1:3} \backslash \mathcal{P}_{D,4} = \{ A, B, C \}$. Note that in Fig.~\ref{fig:frmk}, we neglect the incoming edges from $D$'s non-neighbors into Topo-LSTM unit of $D$, in order to keep the diagram uncluttered.


\stitle{Formulation of Topo-LSTM.}
We now formalize the running example to develop the Topo-LSTM model. For a cascade sequence $s = \{(v_1, 1), ..., (v_T, T)\}$, to infer $v_t$'s embedding $\boldh_t$, we consider three parts of information: (1) $v_t$'s feature vector $\boldx_t$; (2) the possible activation attempts from $v_t$'s precedent set $\mathcal{P}_{v_t, t}$; (3) the other already activated nodes $Q_{1:t-1} \backslash \mathcal{P}_{v_t, t}$. For (2) and (3), because all the nodes in $Q_{1:t-1}$ already have their embedding inferred by time $t$, we can just use their $\boldh_i$'s (for $i=1, ..., t-1$). 

Next we introduce Topo-LSTM, which makes two important changes to the standard LSTM to accommodate the dynamic DAG structure.
\begin{itemize} 
\item \emph{\textbf{Different types of inputs}}: for each node $v_t$, there are two types of active nodes that can contribute to learn $v_t$'s sender embedding, including: 1) the active nodes that are directly linked with $v_t$, denoted as  $v_i \in \mathcal{P}_{v_t, t}$; 2) the other active nodes that are not linked with $v_t$, denoted as $v_j \in (Q_{1:t-1} \backslash \mathcal{P}_{v_t, t})$. These two different types of active nodes are expected to contribute differently to $v_t$'s sender embedding learning, thus we should separate them. Comparatively, in standard LSTM all the cells consider the same type of inputs. 
\item \emph{\textbf{Multiple inputs in each type}}: for each node $v_t$, we have multiple inputs from each type of active nodes, including: 1) the send embedding $\boldh_i$'s for $v_i \in \mathcal{P}_{v_t, t}$; 2) the sender embedding $\boldh_j$'s for $v_j \in (Q_{1:t-1} \backslash \mathcal{P}_{v_t, t})$. Therefore, to compute the contribution of each type of active nodes, we need to aggregate its multiple inputs (either the $\boldh_i$'s or the $\boldh_j$'s). Comparatively, in standard LSTM each cell only takes one input from its precedent node.
\end{itemize}
To incorporate these two important differences, we change the memory cell design of the standard LSTM. 
Specifically, we separate the two types of inputs, and for each type of input, we aggregate the corresponding multiple nodes' embeddings:
\begin{align}
  \boldh_t'^{(p)} &= \phi(\{ \boldh_v | v \in \mathcal{P}_{v_t, t} \}), \\
  \boldh_t'^{(q)} &= \phi(\{ \boldh_v | v \in Q_{1:t-1} \backslash \mathcal{P}_{v_t, t} \}),
\end{align}
where the superscript $^{(p)}$ denotes the input aggregation for the active nodes that are directly linked with $v_t$, and the superscript $^{(q)}$ denotes the input aggregation for the other active nodes that are not linked with $v_t$. Besides, $\phi(\cdot)$ is an aggregation function; \eg, we can define it either as a simple pooling or with a more sophisticated \emph{attention} mechanism \cite{DBLP:conf/icml/XuBKCCSZB15,DBLP:conf/emnlp/LuongPM15}. In this paper, we use \emph{mean pooling} and leave the other kinds of definitions as future work. Similar to LSTM, our Topo-LSTM also has a memory cell with several gates and a cell state. In the following, we denote $d$ as the hidden dimensionality of LSTM, $m=|\mathcal{V}|$, and $\odot$ as an elementwise multiplication. In the following, we continue to use the superscripts $^{(p)}$ and $^{(q)}$, to differentiate the parameters that are used for modeling the inputs from the active nodes linked with $v_t$ or not. 
{\flushleft $\bullet$ \emph{Input gate}}: Denote $\inW \in \mathbb{R}^{d \times m}$, $\inU^{(p)} \in \mathbb{R}^{d \times d}$, $\inU^{(q)} \in \mathbb{R}^{d \times d}$ and $\inb \in \mathbb{R}^d$ as parameters.
We compute the input gate activation vector $\boldi_t \in \mathbb{R}^{d}$ as
\begin{equation}
  \boldi_t = \sigma(\inW ~ \boldx_t + \inU^{(p)} ~ \boldh_t'^{(p)} + \inU^{(q)}~ \boldh_t'^{(q)} + \inb).
\end{equation}
{\flushleft $\bullet$ \emph{Forget gates}}: Denote $\fgetW \in \mathbb{R}^{d \times m}$, $\fgetUp^{(p)} \in \mathbb{R}^{d \times d}$, $\fgetUp^{(q)} \in \mathbb{R}^{d \times d}$, $\fgetUq^{(p)} \in \mathbb{R}^{d \times d}$, $\fgetUq^{(q)} \in \mathbb{R}^{d \times d}$ and $\fgetb \in \mathbb{R}^d$ as parameters. Because we have two different types of inputs, inspired by the Tree-LSTM~\cite{DBLP:conf/acl/TaiSM15}, we introduce two separate forget gates to control how much information we should forget for each type of inputs. 
Then we compute the forget gate activation vectors $\boldf_t^{(p)} \in \mathbb{R}^d$ and $\boldf_t^{(q)} \in \mathbb{R}^d$ as
\begin{align}
  \boldf_t^{(p)} &= \sigma(\fgetW ~ \boldx_t + \fgetUp^{(p)}~ \boldh_t'^{(p)} + \fgetUq^{(p)}~ \boldh_t'^{(q)} + \fgetb), \\
  \boldf_t^{(q)} &= \sigma(\fgetW ~ \boldx_t + \fgetUp^{(q)}~ \boldh_t'^{(p)} + \fgetUq^{(q)}~ \boldh_t'^{(q)} + \fgetb).
\end{align}
{\flushleft $\bullet$ \emph{Cell states}}: To take the cell states from two different types of inputs, we also define two separate aggregation functions:
\begin{align}
  \boldc_t'^{(p)} &= \phi(\{ \boldc_v | v \in \mathcal{P}_{v_t, t} \}), \\
  \boldc_t'^{(q)} &= \phi(\{ \boldc_v | v \in Q_{1:t-1} \backslash \mathcal{P}_{v_t, t} \}). 
\end{align}
Denote $\cellW \in \mathbb{R}^{d \times m}$, $\cellU^{(p)} \in \mathbb{R}^{d \times d}$, $\cellU^{(q)} \in \mathbb{R}^{d \times d}$ and $\cellb \in \mathbb{R}^d$ as parameters. 
We compute the cell activation $\boldc_t \in \mathbb{R}^d$ as
\begin{align}
  \tilde{\boldc}_t &= \tanh(\cellW ~ \boldx_t + \cellU^{(p)} ~ \boldh_t'^{(p)} + \cellU^{(q)} ~ \boldh_t'^{(q)} + \cellb), \\
  \boldc_t &= \boldi_t \odot \tilde{\boldc}_t + \boldf_t^{(p)} \odot \boldc_t'^{(p)} + \boldf_t^{(q)} \odot \boldc_t'^{(q)}.
\end{align}
{\flushleft $\bullet$ \emph{Output gate}}: Denote $\outW \in \mathbb{R}^{d \times m}$, $\outU^{(p)} \in \mathbb{R}^{d \times d}$, $\outU^{(q)} \in \mathbb{R}^{d \times d}$ as weight matrices, $\outb \in \mathbb{R}^d$ as a bias vector. We compute the output gate activation vector $\boldo_t \in \mathbb{R}^{d}$ as
\begin{equation} 
    \boldo_t = \sigma(\outW ~ \boldx_t + \outU^{(p)} ~ \boldh_t'^{(p)} + \outU^{(q)} ~ \boldh_t'^{(q)} + \outb).
\end{equation}
\noindent Finally, the output vector $\boldh_t \in \mathbb{R}^{d}$ at time $t$ is
\begin{equation} \label{eq.topoLSTM.output}
 \boldh_t = \boldo_t \odot \tanh(\boldc_t). 
\end{equation}

\subsection{Activation Prediction} \label{sect.activation_pred}
At time $t+1$, given the sender embeddings $\{\boldh_1, ..., \boldh_t \}$ and the diffusion topology $\mathcal{G}^*_{t+1}$, we can predict an activation probability for each inactive node $v$. We first assume that an activation can happen from an active node to its inactive neighbors. For example, according to the diffusion topology in Fig.~\ref{fig:diffusion_topology}(b), we can activate $C$ by the already activated neighbors $A$ and $B$. Denote $\boldg_v \in \mathbb{R}^{d}$ as the receiver embedding for node $v$ and $b_v \in \mathbb{R}$ as the bias parameter for $v$. We compute the score for activations as
\begin{equation} \label{eq.act_explicit}
   \rho_{v, t+1} = \phi( \{ \boldh_i | v_i \in \mathcal{P}_{v,t+1} \}) \cdot \boldg_v + b_v.
\end{equation}
In practice, there could also exist some node $v$, which becomes activated even though it does not have any edge to the already active nodes in the data graph; \eg, node $D$ in Fig.~\ref{fig:diffusion_topology}(d). This is possible because that some edges between this node $v$ and those already activated nodes are missing in the data graph. Therefore, to address this issue, we further extend Eq.~\ref{eq.act_explicit} to include all the potential interactions between $v$ and all the already active nodes in $Q_{1:t}$:
\begin{equation} \label{eq.act_all}
   \rho_{v, t+1} = \phi( \{ \boldh_i | v_i \in Q_{1:t} \}) \cdot \boldg_v + b_v.
\end{equation}
Once we have computed the score for each inactive node $v \in \mathcal{V} \backslash Q_{1:t}$, we can define the probability of activating $v$ by
\begin{equation} \label{eq.act_prob}
   p(v | \mathcal{G}^*_{t+1}) = \frac{\exp( \rho_{v, t+1} )}{\sum_{u \in \mathcal{V}\backslash Q_{1:t}} \exp(\rho_{u, t+1})}.
\end{equation}

\subsection{Objective Function and Algorithm} 
Our ultimate task is to fit a model $\mathcal{M}$ from the training cascade sequences $\mathcal{S} = \{ s_1, ..., s_n \}$. 
Since we have known how to estimate the activation probability for each node in a cascade at every time step, we can now develop the overall objective function for Topo-LSTM. Denote the $k$-th training cascade as $s_k = \{ (v_{k,1}, 1), ...,  (v_{k, T_k}, T_k) \}$. We want $\mathcal{M}$ to maximize the activation probability at each time step for $v_{k,t}$ $(t=2, ..., T_k)$. 
Thus for $s_k$, we want to maximize
\begin{equation}
   \sum_{t=2}^{T_k} \log p(v_{k,t} \mid \mathcal{G}^*_{k, t}), 
\end{equation}
where $p(v_{k,t} | \mathcal{G}^*_{k,t})$, as defined in Eq.~\ref{eq.act_prob}, relies on the sender embedding computed from Topo-LSTM in Sect.~\ref{sect.dt_embedding}. 
We denote the parameters for sender embedding as $\Theta^{(emb)} = \{ 
\inW, \inU^{(p)}, \inU^{(q)}, \inb, 
\fgetW, \fgetUp^{(p)}, \fgetUq^{(p)}$, $\fgetUp^{(q)}, \fgetUq^{(q)}, \fgetb, 
\cellW, \cellU^{(p)}, \cellU^{(q)}, \cellb, 
\outW, \outU^{(p)}, \outU^{(q)}, \outb \}$, and the parameters for activation prediction as $\Theta^{(act)} = \{\boldg_v, b_v \mid v \in V \}$. 
In all, our target model is characterized by $\mathcal{M} = \{ \Theta^{(emb)}, \Theta^{(act)} \}$. 
Finally, for all the cascade sequences $\mathcal{S}$, we can define the overall objective function to minimize as
\begin{equation} \label{eq.objective}
 \mathcal{L} =  - \frac{1}{\sum_{k=1}^n (T_k - 1)} \sum_{k=1}^{n} \sum_{t=2}^{T_k} \log p(v_{k,t} \mid \mathcal{G}^*_{k, t}) + \lambda \Omega(\mathcal{M}),
\end{equation}
where $\lambda \geq 0$ is a trade-off parameter. $\Omega(\mathcal{M})$ is a regularization function over $\mathcal{M}$; \eg, it sums up the $\ell_2$-norm of each parameter in $\mathcal{M}$.

\stitle{Topo-LSTM Algorithm}. 
We summarize the learning algorithm for Topo-LSTM in Alg.~\ref{alg.topoLSTM-Learn}. 
We use stochastic gradient descent (SGD) for optimization. We first initialize the model parameters $\mathcal{M}$. Then for each training cascade sequence $s_k$, we iterate through each of its nodes $v_{k,t}$'s to construct a diffusion topology. Specifically, we first extract the active nodes in $s_k$ so far by the time $t$ as $Q_{1:t}$ (line 5). After that, we construct the diffusion topology $G^*_{k,t}$ from the data graph $\mathcal{G}$ and $Q_{1:t}$ according to Def.~\ref{def.dt} (line 6). Based on $G^*_{k,t}$, we can compute $\log p(v_{k,t})$ according to Eq.~\ref{eq.act_prob}, and thus the loss. Finally, we compute gradient of $\mathcal{M}$ as $\nabla_\mathcal{M}$ and do gradient descent on $\mathcal{M}$, using for example Adam \cite{kingma2014adam}.

\begin{algorithm}[t]
\caption{Topo-LSTM.Learn}
\label{alg.topoLSTM-Learn}
\begin{algorithmic}[1]
    \REQUIRE Training cascades $\mathcal{S} = \{ s_1, ..., s_n \}$ with each $s_k = \{ (v_{k,1}, t_{k,1}), ...,  (v_{k, T_k}, t_{k, T_k}) \}$, data graph $\mathcal{G}$, parameters $\mathcal{M}$.
   \ENSURE Optimized parameters $\mathcal{M}^*$.
    \STATE Initialize paramters $\mathcal{M}^*$;
	\STATE Training examples $\mathcal{D}\leftarrow \emptyset$;
 	\FOR{$k=1:n$}
		\FOR{$t=2:T_k$}
			\STATE $Q_{1:t-1} \leftarrow $ sequence of active nodes by time $t-1$ in $s_k$;
			\STATE Construct diffusion topology $G^*_{k,t}$ from $\mathcal{G}$ and $Q_{1:t-1}$;
			\STATE $\mathcal{D}\leftarrow \mathcal{D}\cup (G^*_{k,t}, v_{k,t})$;
		\ENDFOR
	\ENDFOR
    \STATE Generate mini-batches $\mathcal{B}$ from $\mathcal{D}$;
    \STATE Use stochastic gradient descent (SGD) to optimize the objective function in Eq.~\ref{eq.objective} w.r.t. $\mathcal{M}$ given $\mathcal{B}$, until convergence. 
\end{algorithmic}
\end{algorithm}

\stitle{Complexity Analysis.} 
We analyze the running complexity of Alg.~\ref{alg.topoLSTM-Learn}. 
Finding $Q_{1:t}$ (line 5) can be done in constant time, since each cascade sequence generally has a limited length. To construct the diffusion topology (line 6), we make use of the monotonically growing property of diffusion topology (as discussed in Sect.~\ref{sect.data_model}) to assist the complexity analysis. In particular, for a cascade sequence $s_k$, we can construct its $G^*_{k,t}$'s gradually, by adding directed edges based on the data graph $\mathcal{G}$. Thus the number of edges in $G^*_{k,T_k}$ is smaller than that of $\mathcal{G}$. This means the complexity is at most $|\mathcal{E}|$. It is worth noting that, although we perform the topology construction again and again for each timestep of each training cascade, in fact these $G^*_{k,t}$'s only need to be computed once. In SGD (line 11), we need to compute the activation probability $p(v_{k,t})$ and the regularization term of $\mathcal{M}$, both of which require a complexity of $O(|\mathcal{V}|)$. To compute the gradient \wrt $p(v_{k,t})$ and $\Omega(\mathcal{M})$, which again require a complexity of $O(|\mathcal{V}|)$. The gradient descent over $\mathcal{M}$ also requires a complexity of $O(|\mathcal{V}|)$. Therefore, we have the overall complexity as $O(|\mathcal{E}| + |\mathcal{V}| \sum_{k=1}^n T_k)$. That is, our algorithm complexity is linear to the data graph size (\ie, $|\mathcal{E}|$ and $|\mathcal{V}|$) and the cascade size (\ie, $\sum_{k=1}^n T_k$).

%% file: experiment_section.tex
\section{Experiments}

\noindent{\bf Datasets.}
We conduct experiments on three public real world datasets. The statistics of the datasets are listed in Table~\ref{tab:data}.

\begin{itemize}
\item {\it Digg}~\cite{hogg2012social} contains diffusions of stories as voted by the users, along with friendship network of the users.
\item {\it Twitter}~\cite{hodas2013simple} contains the diffusion of URLs on Twitter during 2010 and the follower graph of users.
\item {\it Memes}~\cite{leskovec2009meme} contains the diffusion of memes in April 2009 over online news websites; we create a link between two websites if one of them appears earlier than the other in any cascade.
\end{itemize}

\begin{table}[!h]
	\centering
	\caption{Statistics of datasets.}
	\label{tab:data}
	\begin{tabular}{c|ccc}
		\hline
		& Digg      & Twitter   & Memes	\\ \hline
		\# Nodes    & 279,632   & 137,093   & 5,000        \\
		\# Edges    & 2,617,993 & 3,589,811 & 313,669             \\
		\# Cascades & 3,553     & 569    & 54,847       \\
		Avg. cascade length & 30.0 & 5.7 & 17.0 \\
		\hline
	\end{tabular}
\end{table}


For all these data sets, we randomly sample 75\% of all cascades to generate training examples and the rest for testing. We further randomly sample 10\% of the training cascades for validation. \\


\noindent{\bf Baselines.}
We select four state-of-the-art and representative baselines for comparison with our Topo-LSTM model. 
The baselines can be regarded as under two categories: 
1) representation learning methods: Embedded-IC, DeepCas, DeepWalk;
2) non-representation learning methods: IC-SB. 


\begin{itemize}
\item {\it IC-SB}~\cite{DBLP:conf/wsdm/GoyalBL10} infers the diffusion probability $p_{u, v}$ of each edge $(u,v)\in\mathcal{E}$ given training cascades, and predicts diffusion under the classical IC framework, with the probability of activating an inactive node $v$ at time $t$ given by $1-\prod_{u\in\mathcal{P}_{v,t}} (1-p_{u,v})$, where $\mathcal{P}_{v,t}$ is defined in Sect.~\ref{sect.data_model}. We use the Static Bernoulli (SB) in their paper which shows the best performance for our problem setting.
\item {\it Embedded-IC}~\cite{DBLP:conf/wsdm/BourigaultLG16} grounds in the IC framework. It embeds nodes in a latent diffusion space learned from the observed cascades. Then the diffusion probabilities between nodes are computed based on their distances in the embedding space.
\item {\it DeepCas}~\cite{DBLP:conf/www/LiMGM17} represents a diffusion by some sampled paths from the induced diffusion subgraph. A GRU network with an attention mechanism transforms the these paths into a single vector to represent the diffusion. We replace the diffusion size regressor at the end of their pipeline with a logistic classifier to predict node activations.
\item {\it DeepWalk}~\cite{perozzi2014deepwalk} represents the simple baseline which computes the embedding of nodes without using the cascade information, and aggregate the embeddings of the active nodes by mean pooling to represent the diffusion. We then use a logistic classifier to predict diffusion.	
\end{itemize}

We choose the hyperparamters for each baseline as follows. For Topo-LSTM and DeepCas, the hidden dimensionality $d$ is set to 512. For DeepCas, we generate 200 walks of length 10 for each cascade, the same setting as in \cite{DBLP:conf/www/LiMGM17}. For Embedded-IC and DeepWalk, we set $d$ to 64 and 128 respectively, which give the best empirical performance on validation sets.

\stitle{Evaluation metrics.}
Given the current diffusion, predicting the next active node can be viewed as a retrieval problem~\cite{DBLP:conf/wsdm/BourigaultLG16} due to the large number of potential targets. Specifically, the model ranks the inactive nodes by their predicted activation probabilities, and the actual node to be activated next is the (single) relevant item. We regard predicting future activations as a retrival task and use ranking measures due to two main considerations: (1) since each unactivated node could possibly be activated next, there are massive potential targets, thus it is usually unrealistic to predict exactly the next node; (2) it is often useful enough to provide a short list of most likely future activations instead the exact single next node. For evaluation, we use two widely adopted ranking metrics (varying $k$ in $\{10,50,100\}$):

\begin{itemize}
\item {\it Hits@$k$}: The rate of the top-$k$ ranked nodes containing the next active node.
\item {\it MAP@$k$}: The classical Mean Average Precision measure.
\end{itemize}


\begin{table*}[t]
	\centering
	\caption{Results on information diffusion prediction.}
	\label{tab:eval}
	\begin{tabular}{c|ccc|ccc|ccc}
		\hline
		& \multicolumn{3}{c|}{Digg}                          & \multicolumn{3}{c|}{Twitter}                        & \multicolumn{3}{c}{Memes}                          \\ \hline
		MAP@$k$ (\%)& @10         & @50             & @100            & @10             & @50             & @100            & @10             & @50             & @100            \\ \hline
		IC-SB          & 3.624          & 4.584           & 4.800             & 9.268           & 9.819           & 9.834           & 18.220           & 19.428          & 19.558          \\
		DeepWalk    & 3.288          & 4.088           & 4.289           & 15.021          & 15.324          & 15.345          & 13.523          & 14.636          & 14.798          \\
		Embedded-IC & 2.812          & 3.564           & 3.755           & 11.984          & 12.433          & 12.480           & 18.270           & 19.247          & 19.374          \\
		DeepCas & 3.743          & 4.632           & 4.842           & 17.039          & 17.305          & 17.330           & 19.564           & 20.618          & 20.753          \\
		
		Topo-LSTM    & \textbf{5.862} & \textbf{6.842}  & \textbf{7.031}  & \textbf{20.548} & \textbf{20.779} & \textbf{20.805} & \textbf{29.000}     & \textbf{29.933} & \textbf{30.037} \\ \hline
		Hits@$k$ (\%) & @10    & @50             & @100            & @10             & @50             & @100            & @10             & @50             & @100            \\ \hline
		IC-SB          & 10.826         & 33.113          & 48.412          & 22.151          & 31.242          & 32.266          & 41.356          & 65.884          & 74.868          \\
		DeepWalk    & 9.689          & 29.985          & 44.342          & 24.712          & 30.730           & 32.266          & 28.315          & 51.193          & 62.617          \\
		Embedded-IC & 8.887          & 26.117          & 39.220           & 25.134          & \textbf{33.493} & \textbf{36.597} & 35.124          & 55.966          & 65.053          \\
		DeepCas & 10.269          & 30.826          & 45.741           & 25.661          & 31.190 & 33.173 & 38.858          & 60.478          & 69.921          \\		
		Topo-LSTM    & {\bf 15.410}          & \textbf{37.363} & \textbf{50.384} & \textbf{28.279} & 33.152          & 34.897          & \textbf{50.781} & \textbf{69.548} & \textbf{76.850}  \\ \hline		
	\end{tabular}
\end{table*}

\stitle{Comparisons with baselines.}
We compare Topo-LSTM with the baselines on diffusion prediction. As shown in Table~\ref{tab:eval}, the results show an overall trend that the accuracy improves as $k$ increases, as expected since the target is more likely to be included with more candidate nodes retrieved. In comparison with baselines, on the MAP measure, Topo-LSTM improves the best baselines by 20.1\% (Twitter, MAP@100) to 56.6\% (Digg, MAP@10) relatively across all datasets. On the Hits measure, Topo-LSTM improves the best baselines by 2.7\% (Memes, Hits@100) to 42.3\% (Digg, Hits@10) relatively on Digg and Memes. It also improves the best baselines by 10.2\% relatively for Hits@10 on Twitter. These results shows that with explicit modeling of the dynamic DAG structure, Topo-LSTM can better use the topological information of a diffusion than DeepCas. The results also show that, by learning a dynamic sender embedding, Topo-LSTM can better capture the complex diffusion dynamics and interactions among active nodes, which are important to activation prediction, than Embedded-IC, IC-SB, and DeepWalk, which either learn static representations of active nodes or consider the influence of each active node independently. On the other hand, Embedded-IC shows better performance for Hits@$\{50,100\}$ on Twitter. A possible reason is that Twitter has a small number of training cascades; as Embedded-IC has the least number of parameters (node embeddings in a latent diffusion space), it is more robust against overfitting than other methods. In other cases Embedded-IC performs worse, possibly because its latent diffusion space cannot sufficiently capture the complexity of real world diffusions. We also note that the overall performances of all methods are better on the Memes dataset. We could explain this by the fact that the cascades in Memes are mostly short, thus their structure and dynamics are relatively simple for the model to capture.





\begin{figure}[!h]
	\centering
	\includegraphics[width=1\linewidth]{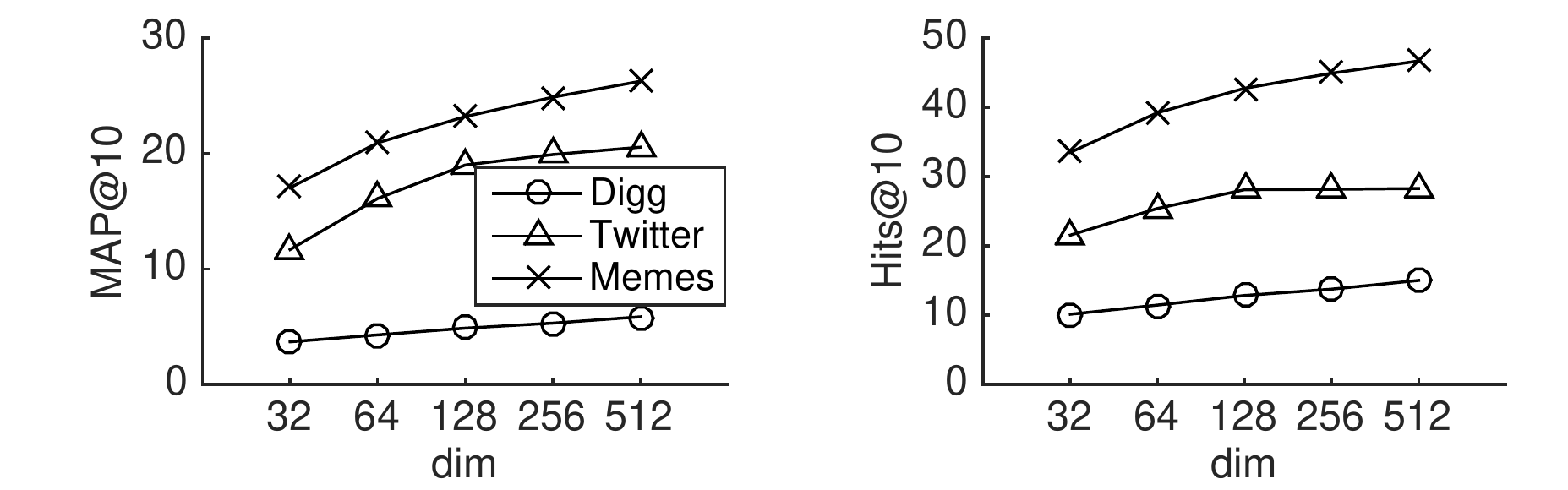}
	\caption{Impact of dimensionality $d$ on Topo-LSTM.}
	\label{fig:dims}
\end{figure}

\stitle{Impacts of hyperparameters.}
We study how the number of hidden dimensions $d$ can affect the performance of Topo-LSTM. As shown in Fig.~\ref{fig:dims}, on Twitter the performance begins to converges at $d$~=~256, possibly due to its small training set, while we could still see steady performance gain on Memes and Digg up to $d$~=~512, indicating that the model has not been saturated; given the large number of observed cascades in Memes and the longer cascade sequences of in Digg, there is room for us to learn with larger models.

\begin{figure*}[t]
	\centering
	\includegraphics[width=1\linewidth]{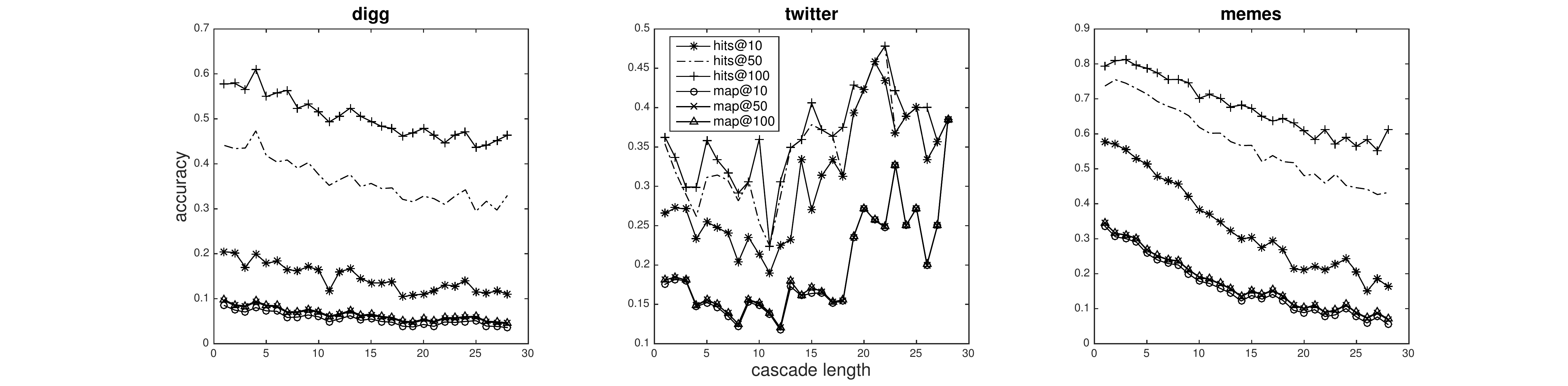}
	\caption{Accuracy of Topo-LSTM at varying cascade length.}
	\label{fig:len}
\end{figure*}

\stitle{Sensitivity to data characteristics.}
We also evaluate how the length of the given cascade could impact the accruacy of predicting future activations. The results are shown in Fig.~\ref{fig:len}. On Digg and Memes, we observe the trend that the prediction accuracy generally decreases as the length of the given cascade increases. That is, it is likely that, in general, future activations are harder to predict for larger cascades. This intuitively makes sense, since given a larger number of active nodes, there are also more potential future nodes to be actived, thus more uncertainty. It is also worth noting that similar phenomena is observed for sentence classification~\cite{DBLP:conf/acl/TaiSM15}, where prediction tasks have better performance on shorter sentences. However, this phenomenon is not observed on Twitter. A possible explanation is that there is higher variation in the propagation paths of tweets, thus future activations for shorter cascades are as well unpredictable as in longer ones.


\begin{table}[!h]
\centering
\caption{Running time (in minutes).}
\label{tab:time}
\begin{tabular}{c|ccc}
\hline
            & Digg & Twitter & Memes \\ \hline
Time (mins) & 140 & 20.8     & 117     \\ \hline
\end{tabular}
\end{table}

\stitle{Running Time.}
We implement Topo-LSTM using Theano 0.9 and conduct all experiments on a machine with an Intel i7-6800K CPU, 32GB memory, and a GTX 1080 GPU. For all the datasets, it takes less than 10 minutes to generate the training examples and less 3 hours to train the model, with our default experiment setting. The detailed running time are reported in Table~\ref{tab:time}. 





%% file: conclusion.tex
\section{Conclusion}
In this paper, we study the problem of predicting future node activations in information diffusion. We adopt a representation learning approach and propose the novel Topo-LSTM model. Tailored for diffusion prediction, Topo-LSTM extends the standard LSTM architecture and is structured as a dynamic DAG. To better model the dynamics of a diffusion, we propose to use the diffusion topology as our new data model, and explicitly model its dynamic DAG structure using Topo-LSTM. As verified by experiments on real world datasets, TopoLSTM improves the best baselines by relatively 20.1\%--56.6\% (MAP) across all the data sets. It also improves the best baselines by relatively 2.7\%--42.3\% (Hits) on both Digg and Memes. Thus, we conclude that our new data model and Topo-LSTM architecture can more effectively capture the diffusion structure as dynamic DAGs.


In the future, we plan to incorporate the content of diffusions and richer node features into our model as additional signals help information diffusion prediction. Besides, we also want to differentiate the importance of each active node in activating a target inactive node, based on how often the active node interacts with the inactive node, when the active node was activated, how long the inactive node has remained unactivated since it was exposed to active neighbors.